\documentclass[runningheads]{llncs}

\usepackage{accv}

\usepackage{accvabbrv}

\usepackage{graphicx}
\usepackage{booktabs}

\usepackage{hyperref}

\usepackage{orcidlink}

\begin{document}

\title{DRIFT: Derailing Denoising Trajectories of Flow-Matching VLAs with Adversarial Patch Attack}
\titlerunning{DRIFT: Derailing Denoising Trajectory of Flow-Matching VLAs}

\author{Hoseong Tae\inst{1,2}\orcidlink{0009-0005-3845-4440} \and
Jong-Seok Lee\inst{1,2}\thanks{Corresponding author}\orcidlink{0000-0002-8038-1119}}

\authorrunning{H. Tae and J.-S. Lee}

\institute{School of Integrated Technology, Yonsei University, Korea \and
BK21 Graduate Program in Intelligent Semiconductor Technology, Yonsei University, Korea \\
\email{\{hoseong.tae, jong-seok.lee\}@yonsei.ac.kr}
}
\maketitle

\begin{abstract}
Flow-matching vision-language-action (VLA) models such as $\pi_0$ generate robot actions by integrating a learned denoising velocity field, and have been reported to resist adversarial perturbations that readily fool autoregressive VLAs.
We show that this robustness is largely illusory: it stems from prior attacks overlooking the multi-step denoising ODE.
We introduce \textbf{DRIFT} (\textbf{D}enoising \textbf{R}edirection via \textbf{I}nput perturbation of the \textbf{F}low-matching \textbf{T}rajectory), a test-time universal adversarial patch placed on the robot's gripper that attacks the denoising velocity field of an off-the-shelf policy. 
Our central finding is counterintuitive: attacking \emph{only the first} denoising step is both stronger and cheaper than attacking a wider window of steps, which we explain through a gradient conflict unique to input-space optimization and which is exactly opposite to the training-time backdoor regime. 
On $\pi_0$ and $\pi_{0.5}$ across four LIBERO suites, DRIFT breaks essentially all originally-solvable tasks with a small single patch, far exceeding action- and embedding-space attack baselines. 
The codes are available at \url{https://taehoseong.github.io/DRIFT/}.
\end{abstract}

\section{Introduction}
\label{sec:intro}

\begin{figure}[t]
\centering
\vspace{-3mm}
\includegraphics[width=\textwidth]{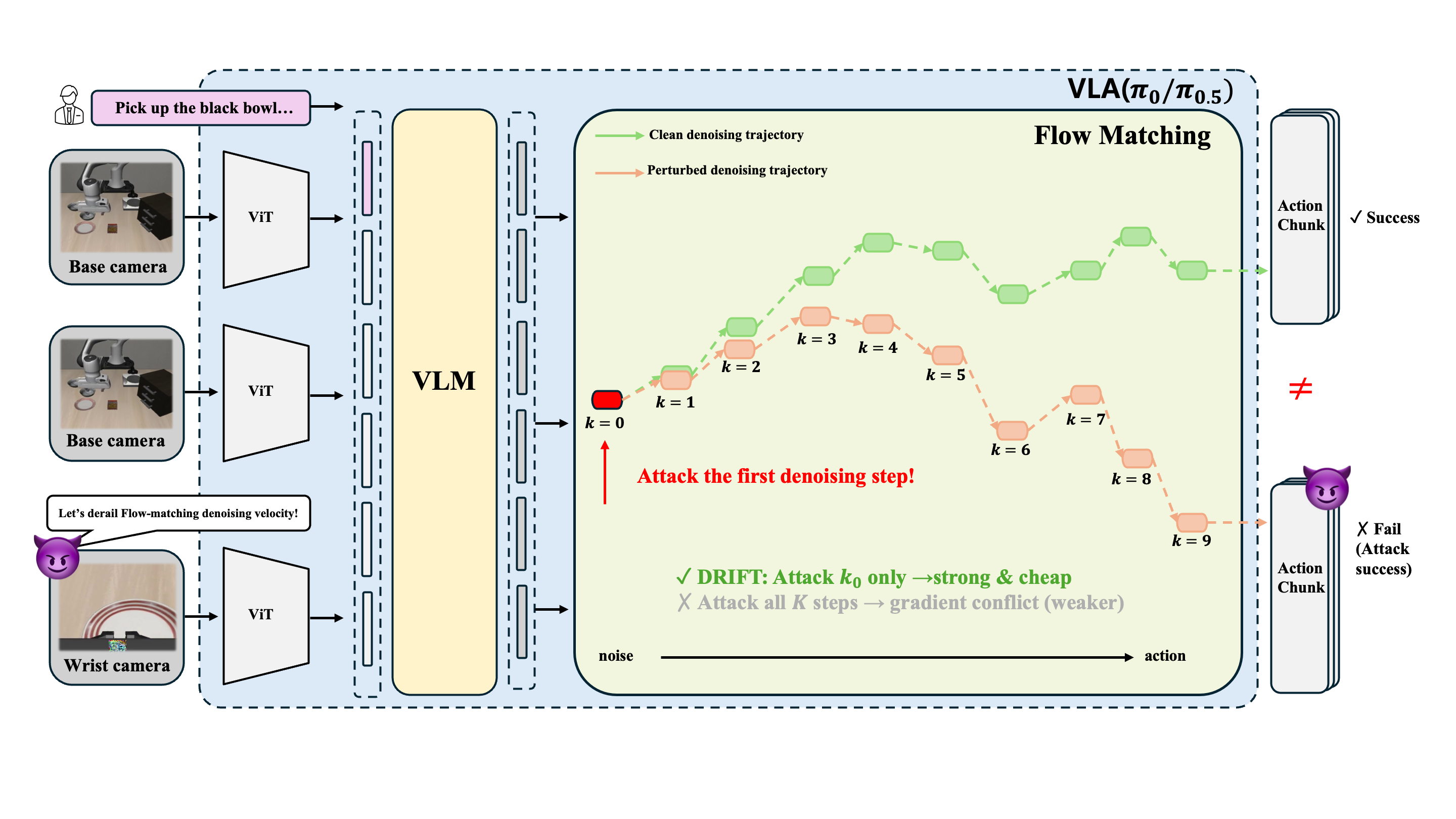}
\vspace{-10mm}
\caption{\textbf{Overview of DRIFT.} A single adversarial patch in the robot's wrist-camera view is optimized once and applied to an off-the-shelf flow-matching VLA ($\pi_0$ or $\pi_{0.5}$) at test time. 
DRIFT perturbs the denoising velocity field at \emph{only the first} step ($k=0$, \textcolor[HTML]{D62728}{red}); the injected error cascades through the denoising ODE, so the perturbed trajectory (\textcolor[HTML]{F57C00}{orange}) progressively diverges from the clean one (\textcolor[HTML]{2E7D32}{green}) over the remaining steps $k=1\!\to\!k=9$ and yields an action chunk that differs from the clean output, turning a successful rollout into a failure.
Attacking $k=0$ alone is both \emph{stronger and cheaper} than attacking the full window of $K$ steps, which is weakened by gradient conflict.}
\label{fig:teaser}
\end{figure}

Vision-Language-Action (VLA) models have emerged as a dominant paradigm for general-purpose robot control, directly mapping raw visual observations and natural-language instructions to low-level actions.
Among them, a family of \emph{flow-matching} VLAs---most prominently $\pi_0$~\cite{black2024pi_0} and its successor $\pi_{0.5}$~\cite{black2025pi05}---has achieved state-of-the-art performance by generating continuous action chunks through an iterative denoising process, rather than autoregressively decoding discrete action tokens.
As these models move toward real-world deployment in safety-critical settings, understanding their adversarial robustness is essential.

A recent benchmark~\cite{guo2026robustness} reports that flow-matching VLAs such as $\pi_0$ are notably robust to naive adversarial perturbations, unlike autoregressive models such as OpenVLA~\cite{kim2025openvla}. 
We argue that this robustness is largely an artifact of how attacks are formulated: current methods perturb the final action space~\cite{wang2025exploring} or the vision-encoder embedding space~\cite{xu2025model} while remaining agnostic to the multi-step denoising ordinary differential equation (ODE) that defines how a flow-matching policy actually produces an action---ignoring the very structure that makes these models most vulnerable.

That the denoising trajectory is an exploitable attack surface is not, by itself, a new observation. 
FlowHijack~\cite{an2026flowhijack} establishes that corrupting the velocity field at the \emph{early}, near-noise phase of integration introduces a small directional error that the ODE solver amplifies along the entire trajectory---an ``early injection, full-path amplification'' effect.
Crucially, however, FlowHijack realizes this through a \emph{training-time backdoor}: it re-trains the policy on poisoned data so that a trigger pattern redirects the learned dynamics at inference. 
This assumes white-box access to the training pipeline and a compromised model, a fundamentally stronger and less practical threat model than perturbing a deployed, frozen policy.
It remains open whether the same cascade vulnerability of an \emph{off-the-shelf} flow-matching VLA can be exploited at test time, with nothing more than a physically realizable adversarial patch.

In this work we answer this question affirmatively, and in doing so reveal that the test-time setting behaves differently from the training-time one. 
We adopt a practical threat model: a single small adversarial patch placed on the robot's gripper, once optimized offline via white-box gradients and applied universally across tasks, requiring no access to model weights at deployment and no modification of the policy. 
Instead of attacking the action output or the visual embedding, we introduce \textbf{DRIFT} (\textbf{D}enoising \textbf{R}edirection via \textbf{I}nput perturbation of the \textbf{F}low-matching \textbf{T}rajectory), which directly attacks the \emph{denoising velocity field}, the quantity that the policy integrates to generate every action (Fig.~\ref{fig:teaser}).

Our central finding concerns not \emph{whether} early steps matter---which prior work has already noted---but \emph{how many} of them to attack. 
Contrary to the natural expectation that perturbing more denoising steps would yield a stronger attack, we find that attacking \textbf{only the first step} is both stronger and cheaper than any wider window; adding more early steps \emph{reduces} success rather than increasing it. 
We trace this ``less-is-more'' effect to a \emph{gradient conflict} specific to input-space optimization (Sec.~\ref{sec:fewer_steps}), and show that it is exactly opposite to the training-time regime, where FlowHijack instead requires a \emph{wide} early window to implant its backdoor---so whether to attack one step or many depends fundamentally on whether one perturbs the input or re\-trains the weights.

We validate these findings on $\pi_0$ and $\pi_{0.5}$ across four LIBERO suites~\cite{liu2023libero}, where DRIFT---our single-step wrist-camera patch---breaks essentially all origin-ally-solvable tasks, far exceeding action-space (UADA~\cite{wang2025exploring}) and embedding-space (EDPA~\cite{xu2025model}) baselines under the same patch-size budget.
We summarize our contributions as follows:

\begin{itemize}
    \item \textbf{A new attack surface and a practical threat model.} We present \textbf{DRIFT}, to our knowledge the first \emph{test-time} adversarial patch attack that explicitly exploits the denoising velocity field of flow-matching VLAs. 
    Unlike FlowHijack~\cite{an2026flowhijack}, which requires a training-time backdoor and a modified model, and unlike action-/embedding-space attacks~\cite{wang2025exploring,xu2025model} that ignore the denoising ODE, DRIFT derails the denoising trajectory of an unmodified, deployed policy through a single physically realizable sticker.

    \item \textbf{Fewer steps, stronger attack.} We show that for a test-time patch, attacking \emph{only the first denoising step} outperforms attacking wider windows of early denoising steps---directly contrasting the training-time backdoor setting, where a narrow early-step window is insufficient.

    \item \textbf{A mechanistic explanation via gradient conflict.} We attribute this ``less-is-more'' effect to a gradient conflict unique to input-space optimization: early- and late-step gradients with respect to the patch are misaligned, so they are canceled when accumulated.
    This analysis, absent from prior training-time work, explains both why the first step suffices and why adding steps hurts.

    \item \textbf{Efficiency.} Because DRIFT requires only a single denoising step in the forward and backward pass, it reduces the optimization cost by a factor of $K$ (the number of denoising steps) relative to full-trajectory attacks, while producing a \emph{stronger} patch---cheaper and more effective.
\end{itemize}
\section{Related Work}
\label{sec:related_work}

\subsection{Vision-Language-Action Models}

VLA models map visual observations and natural-language instructions to robot actions within a single policy, and differ mainly in how they represent actions.
Autoregressive VLAs discretize actions into tokens and decode them like language: RT-1 and RT-2~\cite{brohan2022rt,zitkovich2023rt} combine web-scale vision-language pretraining with robot control, and OpenVLA~\cite{kim2025openvla} provides a popular open model in this family.
Continuous VLAs instead generate real-valued actions directly, better matching the continuous nature of control.
Diffusion Policy~\cite{chi2025diffusion} models visuomotor control as conditional denoising over action trajectories, and Octo~\cite{team2024octo} scales this idea to a generalist multi-embodiment policy. Most relevant to us, $\pi_0$~\cite{black2024pi_0} and its successor $\pi_{0.5}$~\cite{black2025pi05} adopt flow matching to produce action chunks through an iterative denoising ODE---the distinctive structure our attack exploits, and a likely reason attacks designed for autoregressive VLAs do not transfer to them directly.

\subsection{Adversarial Attacks on VLA Models}

Guo~\etal~\cite{guo2026robustness} benchmark VLA robustness and find flow-matching models---notably $\pi_0$---substantially more robust to naive perturbations than autoregressive ones; our work revisits this observation by targeting the denoising trajectory.
Earlier robot-policy attacks perturb visual observations to degrade performance~\cite{lu2026robots,yan2025alignment,huang2026trap,jones2025adversarial}, and DP-Attacker~\cite{chen2024diffusion} shows that diffusion-based visuomotor policies are vulnerable to perturbations crafted against the denoising objective, revealing exploitable structure in the iterative generation process.

More recent work has extended adversarial attacks directly to VLA models. 
UADA~\cite{wang2025exploring} proposes a universal adversarial patch optimized to maximize task failure across diverse manipulation scenarios. 
By treating the patch as a physical sticker placed in the robot's camera view, this approach constitutes a practical test-time threat model that requires no access to the model's training procedure. 
EDPA~\cite{xu2025model} further investigates model-agnostic adversarial attacks on VLA models, showing that perturbations transfer across architectures with different action generation mechanisms, underscoring the broader vulnerability of VLA policies to input-level manipulation.

Backdoor attacks form another line of adversarial research on VLAs~\cite{zhou2026badvla, zhou2025goal, xu2025tabvla}, in which a trigger implanted during training elicits attacker-chosen behavior at inference.
These attacks are largely agnostic to how actions are generated. 
In contrast, FlowHijack~\cite{an2026flowhijack} specifically targets flow-matching VLAs, injecting a backdoor during training by conditioning the denoising trajectory on a trigger pattern so as to steer the robot toward a target behavior. 
While FlowHijack shares our insight that the denoising trajectory is a key attack surface, it operates under a fundamentally different threat model: it requires access to the training pipeline, whereas our work focuses on test-time universal adversarial patches that require no model modification. 
Our analysis further reveals \emph{which} denoising steps are most vulnerable, providing a principled basis for efficient patch optimization.

\section{Preliminary}
\label{sec:preliminary}
\vspace{-2mm}
\subsection{Flow-Matching-Based Action Generation}
\label{sec:flow_matching}
\vspace{-2mm}
We target Vision-Language-Action (VLA) models that generate actions through
flow matching, exemplified by $\pi_0$ and $~\pi_{0.5}$~\cite{black2024pi_0, black2025pi05}. 

Given an observation context $\mathbf{o}_t$ at time $t$, the policy models the conditional distribution $p(\mathbf{A}_t \mid \mathbf{o}_t)$ over a future action chunk
$
    \mathbf{A}_t =
    \left[
    \mathbf{a}_t, \mathbf{a}_{t+1}, \dots, \mathbf{a}_{t+H-1}
    \right],
$
where $H$ denotes the action horizon and each $\mathbf{a}_{t+j}$ is a low-level action command. 
The conditioning observation is composed of multi-view images, a language instruction, and the robot proprioceptive state:
$
    \mathbf{o}_t =
    \left[
    \mathbf{I}_t^1, \dots, \mathbf{I}_t^n, \ell_t, \mathbf{q}_t
    \right],
$
where $\mathbf{I}_t^i$ is the image from the $i$-th camera view, $\ell_t$ the tokenized language instruction, and $\mathbf{q}_t$ the robot proprioceptive state.

Flow matching constructs a continuous interpolation between Gaussian noise and the clean action chunk. 
Specifically, for a sampled action $\mathbf{A}_t \sim p(\mathbf{A}_t \mid \mathbf{o}_t)$ and noise $\boldsymbol{\epsilon} \sim \mathcal{N}(\mathbf{0}, \mathbf{I})$, the noisy intermediate action at flow time $\tau$ is defined as
\begin{equation}
    \mathbf{A}_t^\tau
    =
    \tau \mathbf{A}_t
    +
    (1-\tau)\boldsymbol{\epsilon},
    \quad \tau \in [0,1].
\end{equation}
Equivalently, this corresponds to the conditional distribution
\begin{equation}
    q(\mathbf{A}_t^\tau \mid \mathbf{A}_t)
    =
    \mathcal{N}
    \left(
    \tau \mathbf{A}_t,
    (1-\tau)^2\mathbb{I}
    \right).
\end{equation}
Under this interpolation, the target velocity field is given by
\begin{equation}
    \mathbf{u}(\mathbf{A}_t^\tau \mid \mathbf{A}_t)
    =
    \mathbf{A}_t - \boldsymbol{\epsilon}.
\end{equation}

During training, the model learns a velocity field $\mathbf{v}_\theta(\mathbf{A}_t^\tau, \mathbf{o}_t)$ that predicts the direction from the noisy intermediate action toward the clean action.
The flow-matching training objective is
\begin{equation}
    \mathcal{L}^{\tau}_{\mathrm{train}}(\theta)
    =
    \mathbb{E}_{p(\mathbf{A}_t \mid \mathbf{o}_t),\,
    q(\mathbf{A}_t^\tau \mid \mathbf{A}_t)}
    \left[
    \left\|
    \mathbf{v}_\theta(\mathbf{A}_t^\tau, \mathbf{o}_t)
    -
    \mathbf{u}(\mathbf{A}_t^\tau \mid \mathbf{A}_t)
    \right\|_2^2
    \right].
\end{equation}

At inference time, the policy starts from pure Gaussian noise $\mathbf{A}_t^0 = \boldsymbol{\epsilon} \sim \mathcal{N}(\mathbf{0}, \mathbb{I})$ and iteratively integrates the learned velocity field using $K$ discrete Euler steps with step size $\Delta\tau = 1/K > 0$:
\begin{equation}
    \mathbf{A}_t^{\tau + \Delta\tau}
    =
    \mathbf{A}_t^{\tau}
    +
    \Delta\tau\,
    \mathbf{v}_\theta(\mathbf{A}_t^\tau, \mathbf{o}_t),
    \quad \tau = 0,\,\Delta\tau,\,2\Delta\tau,\,\dots,\,1-\Delta\tau.
\end{equation}
After $K$ steps, the trajectory reaches $\mathbf{A}_t^1 \approx \mathbf{A}_t$, which is returned as the predicted action chunk.
This continuous action-generation process distinguishes flow-matching
VLAs from autoregressive policies that decode discrete action tokens, and it is the main class of models targeted in our analysis.
\vspace{-4mm}
\subsection{Threat Model}
\label{sec:threat_model}
\vspace{-2mm}
\subsubsection{Attack Goal.}
The adversary seeks to induce \emph{task failure}: under the presence of the perturbation, the policy should produce actions that prevent the robot from completing its instructed manipulation task. 
We consider the \emph{untargeted} setting, in which the objective is simply to maximize the failure rate rather than to steer the robot toward a specific attacker-chosen behavior (\eg, a fixed pose). 
This is a strictly weaker assumption than targeted backdoor attacks such as FlowHijack~\cite{an2026flowhijack}, and it directly reflects the safety-critical concern that a deployed robot can be made unreliable by a single physical artifact.
Concretely, given a clean observation $\mathbf{o}_t$ and its perturbed counterpart $\mathbf{o}_t^{+}$, the adversary aims to drive the generated action chunk $\mathbf{A}_t$ away from the behavior the policy would have produced on $\mathbf{o}_t$, thereby lowering the task success rate. 
Since a flow-matching policy realizes $\mathbf{A}_t$ only through the integrated denoising velocity field $\mathbf{v}_\theta$, we operationalize this goal as maximizing the divergence of $\mathbf{v}_\theta$ between the clean and perturbed conditioning, rather than attacking the final action directly (Sec.~\ref{sec:objective}).
\begin{figure}[t]
\centering
    \begin{subfigure}{0.49\textwidth}\centering\includegraphics[width=\linewidth]{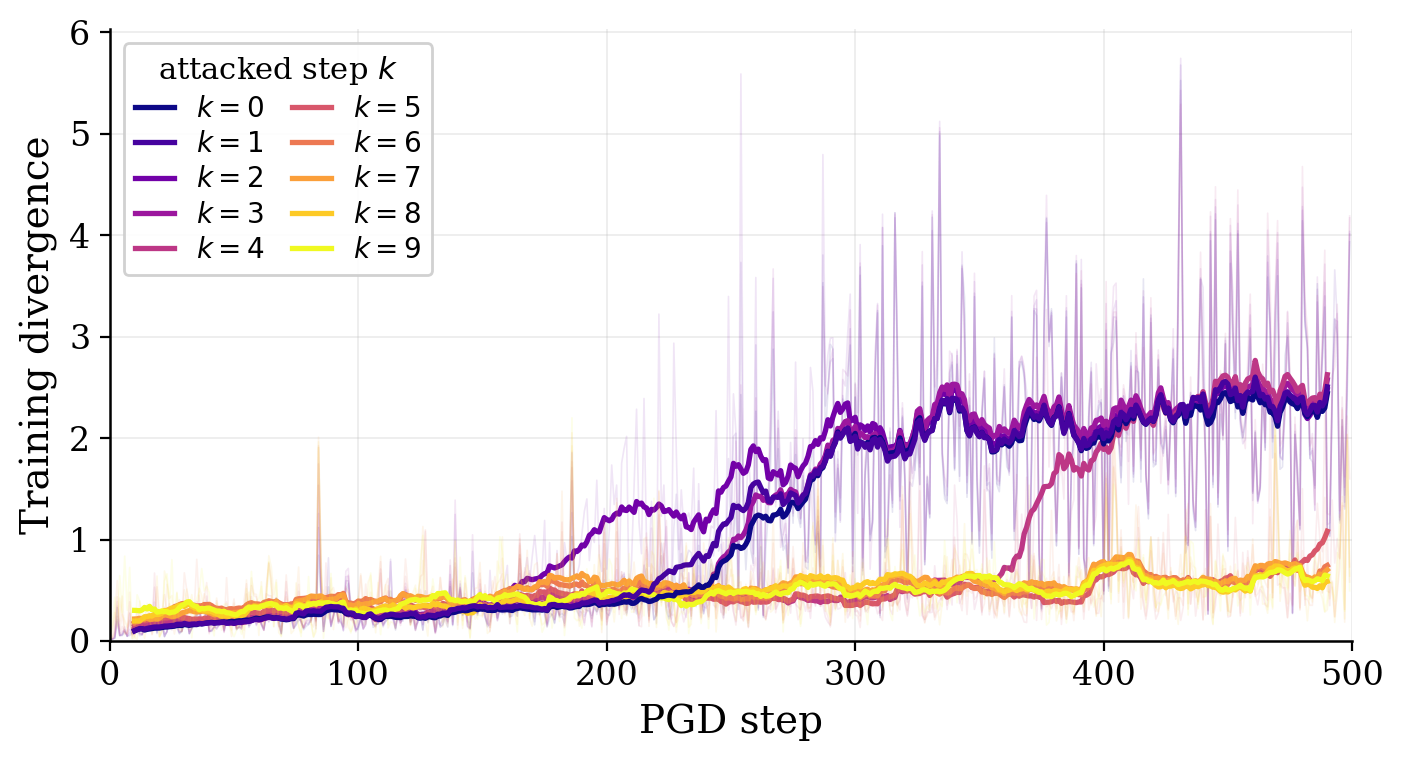}\caption{}\label{fig:single_step_a}\end{subfigure}\hfill
    \begin{subfigure}{0.49\textwidth}\centering\includegraphics[width=\linewidth]{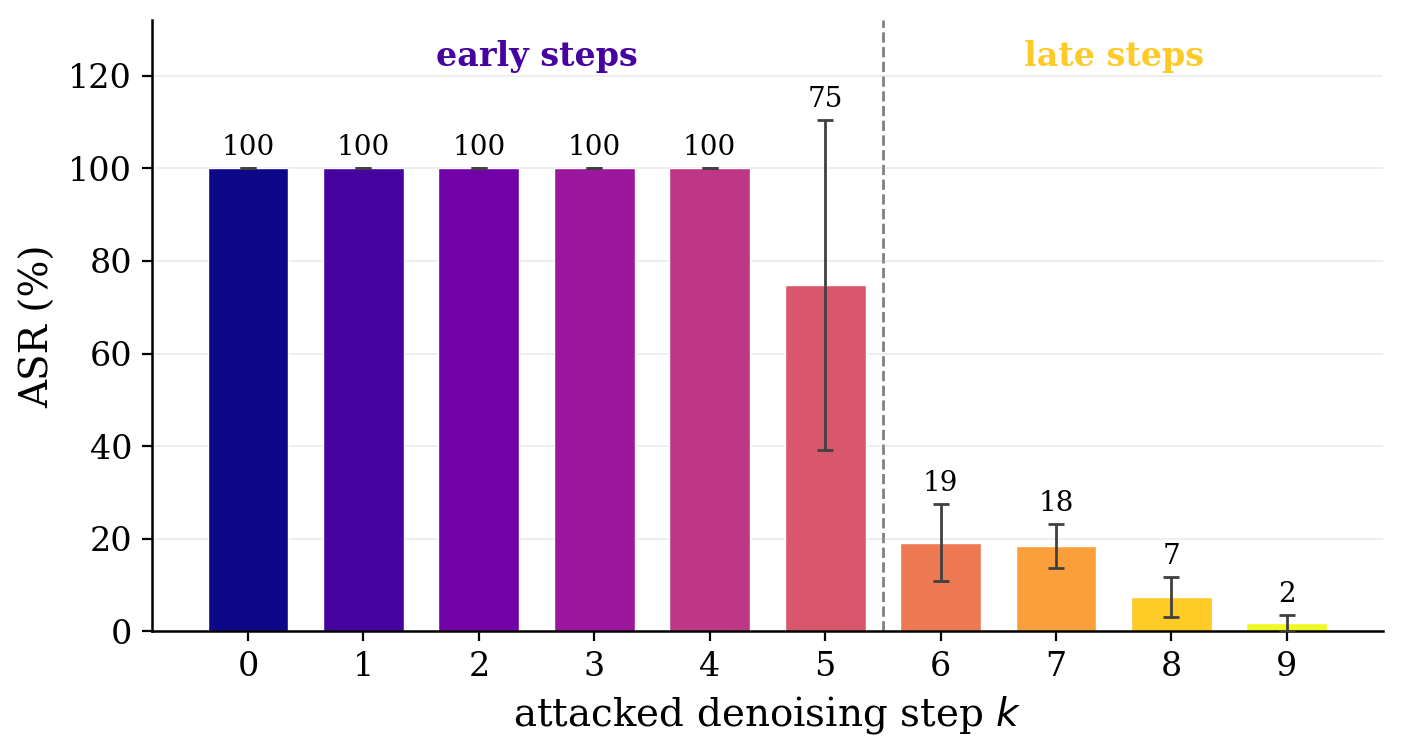}\caption{}\label{fig:single_step_b}\end{subfigure}
    \caption{\textbf{Single-step vulnerability by denoising step $k$.}
    (a)~Training velocity divergence with respect to the step of patch optimization using PGD when only step $k$ is attacked (color: early\,$\rightarrow$\,late); the earliest steps reach the highest divergence.
    (b)~Corresponding ASR on $\pi_0$ (bars: mean over four LIBERO~\cite{liu2023libero} suites; whiskers: $\pm$std, 3 seeds): early steps ($k\le5$) break essentially all tasks, late steps ($k\ge6$) collapse.
    }
    \label{fig:single_step}
    \vspace{-5mm}
\end{figure}
\vspace{-10mm}
\subsubsection{Adversary Capability.}
We adopt the standard white-box patch-attack setting~\cite{wang2025exploring}, but restrict the adversary to test-time, input-space manipulation only. 
Specifically, the adversary:
\begin{itemize}
    \item \textbf{has white-box access to the frozen policy} (architecture and weights) for offline patch optimization, allowing gradients to flow through the denoising process, but \textbf{cannot modify the weights, fine-tune, or poison the training data}. 
    This is the key distinction from training-time backdoor attacks~\cite{an2026flowhijack}, which require a compromised model.
    \item \textbf{can place a single physical patch within the camera's field of view.} Following the practical sticker threat model, the patch occupies a small contiguous image region (32\,px, ${\approx}2$--3\,cm physically, roughly $2\%$ of the wrist image) over the gripper, and only the pixels inside this region are modified; the rest of the observation, the proprioceptive state, and the language instruction remain untouched.
    \item \textbf{deploys a universal patch.} A single patch is optimized offline and then fixed: it is applied unchanged across all tasks, scenes, and episodes, requiring no per-instance optimization and no access to the model at deployment time.
\end{itemize}
This capability profile is intentionally conservative---it assumes no control over the model pipeline and no online feedback---yet, as we show, it is sufficient to reliably break an off-the-shelf flow-matching VLA.
\vspace{-3mm}
\section{Methodology}
\label{sec:methodology}

We first present a controlled analysis that identifies \emph{which} part of the denoising process is most susceptible to an input-space perturbation (Sec.~\ref{sec:analysis}), and then formalize the attack that exploits it (Secs.~\ref{sec:pgd}--\ref{sec:objective}). 
Throughout, we craft a single universal patch $\boldsymbol{\delta}\in[0,1]^{h\times w\times 3}$ that, when pasted onto a fixed region $\mathcal{R}$ of the wrist image, perturbs the conditioning observation from $\mathbf{o}_t$ to $\mathbf{o}_t^{+}=\mathcal{A}(\mathbf{o}_t,\boldsymbol{\delta})$, where the overlay operator $\mathcal{A}$ replaces the pixels in $\mathcal{R}$ with $\boldsymbol{\delta}$ and leaves the remaining observation untouched.

\subsection{Step-wise Vulnerability Analysis}
\label{sec:analysis}

Recall from Sec.~\ref{sec:flow_matching} that a flow-matching policy produces an action chunk by integrating its velocity field $\mathbf{v}_\theta$ over $K$ discrete Euler steps. 
We index these integration steps by $k\in\{0,\dots,K-1\}$, where $k{=}0$ is the first step that departs from pure noise and $k{=}K{-}1$ is the last step adjacent to the final action. 
To attribute the attack's effect to specific steps, we optimize patches against the velocity-divergence objective introduced below while varying \emph{only} the set of attacked steps $\mathcal{S}\subseteq\{0,\dots,K-1\}$ and holding all other hyperparameters fixed. 
This isolates the role of step selection from confounds such as patch size, position, or optimization budget.

\subsubsection{Single-Step Ablation.}
\label{sec:ablation_single}

We begin with the most restrictive setting, $\mathcal{S}=\{k\}$, in which the patch is optimized to corrupt the velocity at a \emph{single} step $k$:
\begin{equation}
    \mathcal{L}_{k}(\boldsymbol{\delta})
    =
    \mathbb{E}_{\mathbf{o}\sim\mathcal{D}}
    \big\|
        \mathbf{v}_\theta(\mathbf{A}^{\tau(k)}_{adv},\mathbf{o}^{+})
        -
        \mathbf{v}_\theta(\mathbf{A}^{\tau(k)}_{clean},\mathbf{o})
    \big\|_2^2 ,
    \label{eq:single_step}
\end{equation}
where $\tau(k)$ denotes $k/K$, $\mathbf{o}^{+}=\mathcal{A}(\mathbf{o},\boldsymbol{\delta})$ is the patched observation and the expectation is taken over $\mathcal{D}$, a pool of collected wrist-camera observations. 
Fig.~\ref{fig:single_step}(a) tracks the training-time divergence of velocity achieved at each $k$: patches that target the earliest steps (small $k$, near the noise end) attain markedly higher divergence than patches that target late steps. 
Crucially, this training signal translates into task failure. 
Fig.~\ref{fig:single_step}(b) reports the resulting attack success rate (ASR): attacking any of the early steps ($k\le5$) breaks essentially all tasks, whereas attacking a late step ($k\ge6$) collapses to near-zero ASR. 
Divergence and ASR agree on the same early/late boundary near $k{=}5$. 
The vulnerability of a flow-matching VLA to an input-space perturbation is therefore \emph{not} uniformly distributed along the denoising path but concentrated at its onset.

Although attacks on any of the first few steps all saturate near $100\%$ ASR under our default attack configuration (leaving them empirically indistinguishable), we target the \emph{first} step for two reasons: it maximizes the cascade horizon (an error at $k{=}0$ is amplified over all remaining $K{-}1$ steps), and it is the \emph{cheapest} single step to attack, since reaching step $k$ requires integrating the ODE through the preceding $k$ steps whereas $k{=}0$ needs no rollout (Sec.~\ref{sec:objective}).

\begin{figure}[t]
\centering

\begin{minipage}[t]{0.38\textwidth}
\vspace{0pt}
\centering

\includegraphics[width=\linewidth]{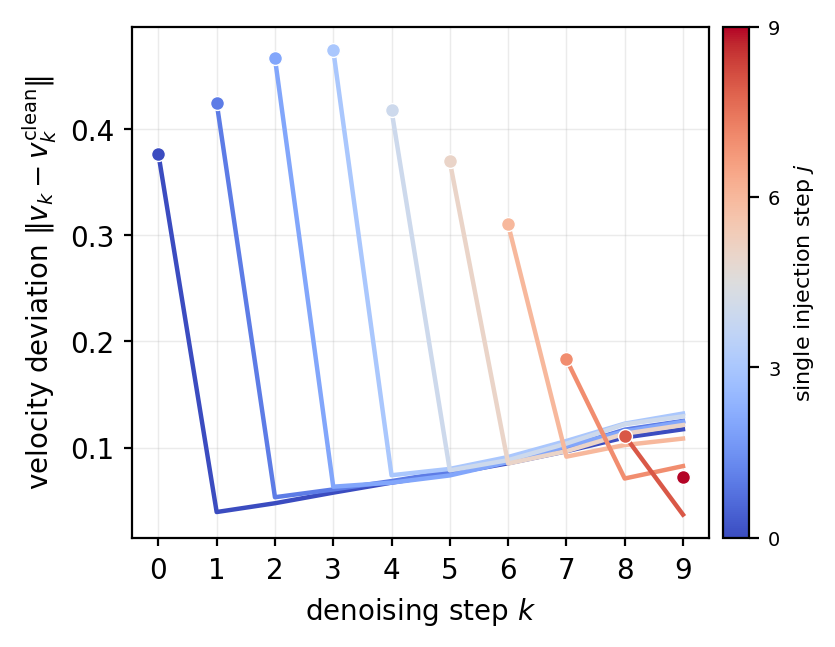}
\vspace{-7mm}
\captionof{figure}{
  \textbf{Velocity Deviation}
}
\label{fig:amplification}
\end{minipage}
\hfill
\begin{minipage}[t]{0.60\textwidth}
\vspace{0pt}
\centering
\vspace{-3mm}
\captionof{table}{
  \textbf{Fewer steps, stronger attack.}
  ASR (\%) of a first-$M$ attack perturbing the $M$ earliest denoising steps
  on $\pi_0$ across four LIBERO suites
  (mean\,$\pm$\,std over 3 seeds).
  Attacking only the first step ($M{=}1$, \ie\ DRIFT) is both strongest and most stable,
  despite using the fewest steps.
}
\label{tab:topk_ablation}

\vspace{2mm}

\setlength{\tabcolsep}{3pt}
\resizebox{\linewidth}{!}{%
\begin{tabular}{lccccc}
\toprule
\textbf{$M$ (attacked steps)}
& \textbf{Spatial}
& \textbf{Goal}
& \textbf{Object}
& \textbf{Long}
& \textbf{Avg} \\
\midrule
\textbf{1} \textit{(DRIFT)}
& \textbf{100.0}\,$\pm$\,0.0
& \textbf{99.7}\,$\pm$\,0.5
& \textbf{100.0}\,$\pm$\,0.0
& \textbf{99.6}\,$\pm$\,0.5
& \textbf{99.8} \\
3
& 73.8\,$\pm$\,12.7
& 59.4\,$\pm$\,27.7
& 90.3\,$\pm$\,8.7
& 86.5\,$\pm$\,9.6
& 77.5 \\
5
& 72.3\,$\pm$\,6.9
& 67.0\,$\pm$\,19.3
& 95.3\,$\pm$\,5.2
& 84.1\,$\pm$\,14.3
& 79.7 \\
\bottomrule
\end{tabular}%
}

\end{minipage}
\vspace{-6mm}
\end{figure}

\subsubsection{Cascade Effect in Early Denoising Steps.}
\label{sec:cascade}
The reason early steps dominate is the autoregressive nature of the ODE solver: a directional error introduced at step $k$ enters the state $\mathbf{A}_t^{\tau}$ and is carried forward through every subsequent Euler update, so its effect compounds along the remaining trajectory.
This ``early injection, full-path amplification'' property of flow-matching dynamics was first articulated by FlowHijack~\cite{an2026flowhijack} in the context of a training-time backdoor.

To show cascade explicitly, we conduct a single-step injection experiment. We inject the adversarial patch at only one denoising step $j$, while using clean visual conditioning at all other steps, and track the resulting velocity deviation throughout the remaining denoising trajectory.
As shown in Fig.~\ref{fig:amplification}, a perturbation injected at the early step initially attenuates but subsequently re-amplifies $\sim3\times$ despite clean conditioning thereafter.
Notably, the instantaneous injected deviation peaks at $k{=}2$--$3$, not $k{=}0$; yet later injections do not exhibit stronger cascade amplification. This explains why optimizing the first step alone is sufficient and effective for DRIFT while requiring the lowest computational cost.

\subsubsection{Fewer Steps, Stronger Attack.}
\label{sec:fewer_steps}
\begin{figure}[t]
\centering
\vspace{-3mm}
\includegraphics[width=0.85\linewidth]{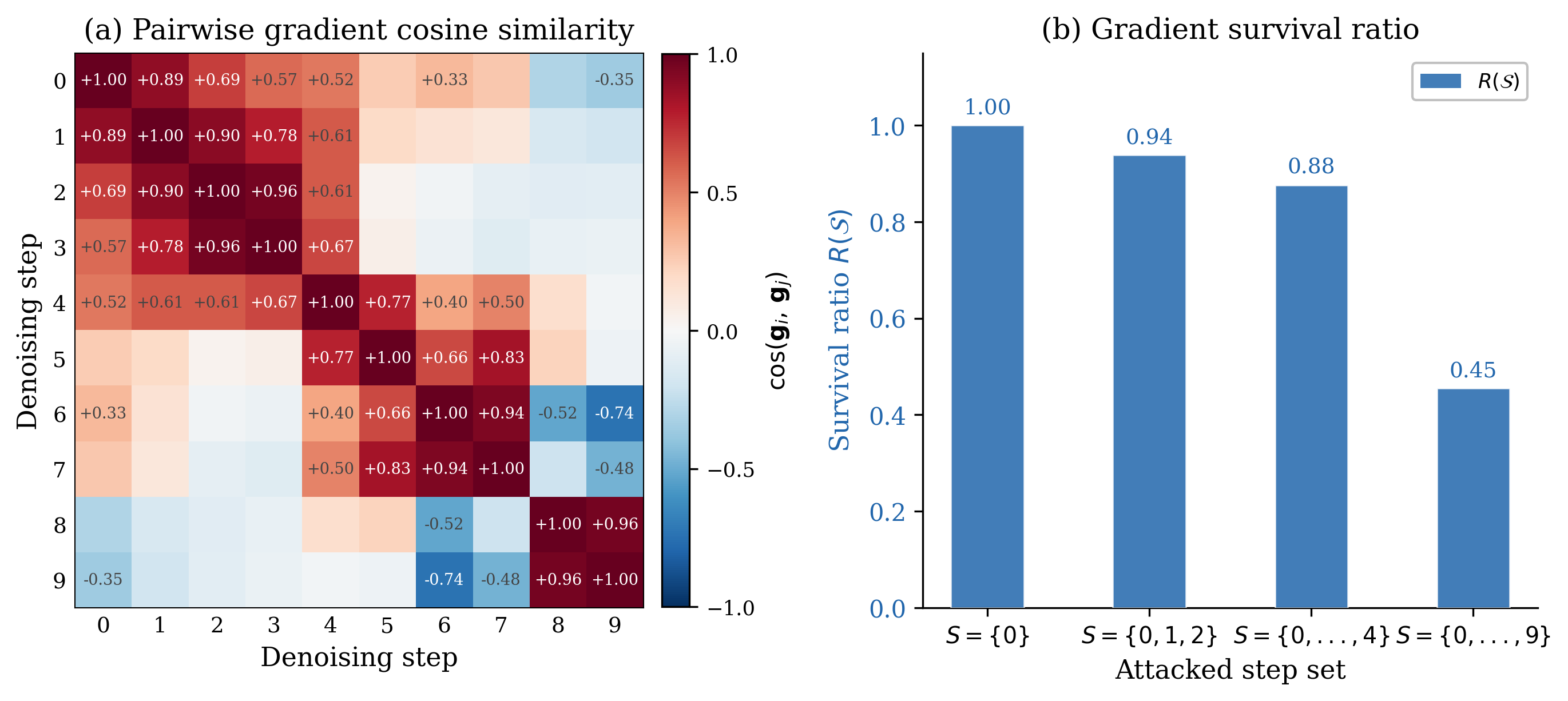}
\vspace{-3mm}
\caption{\textbf{Gradient conflict.}
Left: pairwise cosine similarity between the per-step patch gradients $\mathbf{g}_k=\partial\|\mathbf{v}_\theta(\mathbf{A}^{\tau(k)},\mathbf{o}^{+})-\mathbf{v}_\theta(\mathbf{A}^{\tau(k)},\mathbf{o})\|_2^2/\partial\boldsymbol{\delta}$; the first-step gradient is anti-aligned with the late-step ones (e.g.,\ $\cos(\mathbf{g}_0,\mathbf{g}_9){\approx}{-}0.35$).
Right: for a first-$M$ window we report the survival ratio $R(\mathcal{S})=\|\sum_{k\in\mathcal{S}}\mathbf{g}_k\|\,/\,\sum_{k\in\mathcal{S}}\|\mathbf{g}_k\|$ ($R{=}1$ when all step gradients align, $R{\to}0$ under complete cancellation). 
$R$ falls as the window widens ($1.00{\to}0.45$ from $M{=}1$ to $M{=}10$)---the gradients increasingly cancel---mirroring the measured drop in ASR, so concentrating the attack on the first step alone is strongest.}
\vspace{-5mm}
\label{fig:gradient_conflict}
\end{figure}

Given that early steps cascade, a natural expectation is that attacking a \emph{window} of early steps would be at least as strong as attacking the first one. Surprisingly, we find the opposite. 
Table~\ref{tab:topk_ablation} reports a first-$M$ ablation in which $\mathcal{S}=\{0,\dots,M-1\}$ targets the $M$ earliest steps, summing the single-step divergence of Eq.~\eqref{eq:single_step} over the window:
\begin{equation}
    \mathcal{L}_{\mathcal{S}}(\boldsymbol{\delta})
    =
    \mathbb{E}_{\mathbf{o}\sim\mathcal{D}}
    \sum_{k\in\mathcal{S}}
    \big\|
        \mathbf{v}_\theta(\mathbf{A}^{\tau(k)}_{adv},\mathbf{o}^{+})
        -
        \mathbf{v}_\theta(\mathbf{A}^{\tau(k)}_{clean},\mathbf{o})
    \big\|_2^2 .
    \label{eq:objective}
\end{equation}
The single-step attack ($M{=}1$) is the strongest on average, and \emph{adding} steps degrades the attack. 
We attribute this ``less-is-more'' effect to a \emph{gradient conflict} intrinsic to input-space optimization.
The total objective is a sum of per-step velocity-divergence terms, so the patch gradient is the corresponding sum of per-step gradients $\mathbf{g}_k$. 
As visualized in Fig.~\ref{fig:gradient_conflict}, these per-step gradients are \emph{mutually misaligned}: the first-step gradient $\mathbf{g}_0$ stays well aligned with its immediate neighbors but turns \emph{anti-aligned} with the latest steps (i.e.,\ $\cos(\mathbf{g}_0,\mathbf{g}_9){\approx}{-}0.35$), and the misalignment grows with the distance between steps. 
Summing such conflicting gradients partially cancels the signal and yields a weaker patch (see the supplementary material~\ref{sec:supp_pcgrad} for further validation).

This stands in direct contrast to the training-time backdoor regime. FlowHijack\cite{an2026flowhijack} must inject its malicious dynamics over a \emph{wide} early window---its attack success collapses when the injection window is narrowed toward pure noise, because too narrow a window provides insufficient gradient to \emph{learn} a consistent malicious vector field in weight space.
For a test-time patch on a \emph{frozen} model, there is no such field to learn; the optimization instead seeks a single input perturbation, for which concentrating the entire attack on the first step is sufficient while incurring the lowest computational cost.

\vspace{-4mm}
\subsection{PGD on Denoising Velocity Vector Field}
\label{sec:pgd}
\vspace{-2mm}
Motivated by the analysis above, our attack optimizes the patch to maximize the discrepancy between the clean and perturbed velocity fields at the targeted steps. 
For an observation $\mathbf{o}\in\mathcal{D}$ (the universal observation set of Eq.~\eqref{eq:single_step}) we roll out the denoising trajectory and, at each attacked step $k\in\mathcal{S}$, evaluate the velocity under the clean and perturbed conditioning, $\mathbf{v}_\theta(\mathbf{A}^{\tau(k)},\mathbf{o})$ and $\mathbf{v}_\theta(\mathbf{A}^{\tau(k)},\mathbf{o}^{+})$. The rollout is fully differentiable, so the gradient with respect to $\boldsymbol{\delta}$ propagates through both each step's conditioning and the accumulated adversarial state---that is, a cascade-inclusive gradient. 
The objective sums these per-step gradients, which is precisely the setting in which the gradient conflict of Sec.~\ref{sec:fewer_steps} arises.

We optimize $\boldsymbol{\delta}$ with projected gradient descent (PGD). At each iteration we take a sign-gradient ascent step on the objective $\mathcal{L}$ and project the patch back to the valid image range,
\begin{equation}
    \boldsymbol{\delta} \;\leftarrow\;
    \mathrm{clip}_{[0,1]}\!\Big(
        \boldsymbol{\delta} + \alpha \,\mathrm{sign}\big(\nabla_{\boldsymbol{\delta}}\,\mathcal{L}(\boldsymbol{\delta})\big)
    \Big),
\end{equation}
where $\alpha$ is the step size.
When a stealth budget is desired, we additionally project $\boldsymbol{\delta}$ onto an $\ell_\infty$ ball of radius $\epsilon$ around an initialization image.
Because the patch occupies only the fixed wrist region $\mathcal{R}$ and is shared across all of $\mathcal{D}$, the resulting $\boldsymbol{\delta}$ is a single universal sticker that requires no per-instance optimization at deployment.

\subsection{Training Objective}
\label{sec:objective}

The single-step objective of Eq.~\eqref{eq:single_step} and its first-$M$ window generalization of Eq.~\eqref{eq:objective} together define a family of attacks parameterized by which steps are attacked.
The step-wise analysis of Sec.~\ref{sec:analysis} shows that the most effective member of this family is also the simplest: the single-step attack at the first denoising step, $\mathcal{S}=\{0\}$. This defines our attack, \textbf{DRIFT}, whose objective is $\mathcal{L}_{\text{DRIFT}}(\boldsymbol{\delta})=\mathcal{L}_{0}(\boldsymbol{\delta})$---Eq.~\eqref{eq:single_step} evaluated at $k{=}0$.
This is the objective we use throughout our main experiments. 
It avoids the gradient conflict incurred by attacking additional steps (Sec.~\ref{sec:fewer_steps}), and it requires only a single forward and backward pass through the velocity field per iteration, reducing the optimization cost by a factor of $K$ relative to a full-trajectory attack.
\vspace{-3mm}

\begin{figure*}[t]
\centering
\vspace{-3mm}
\includegraphics[width=\textwidth]{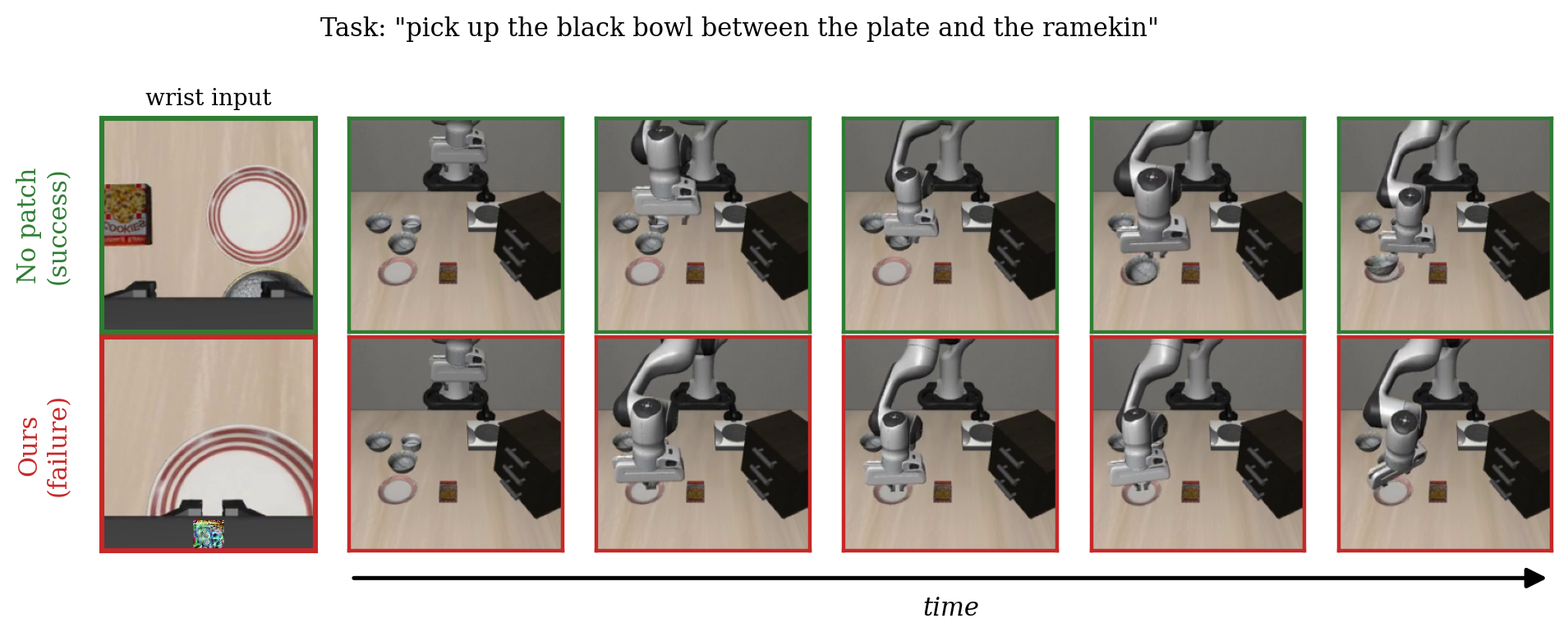}
\vspace{-7mm}
\caption{
  \textbf{Qualitative result on $\pi_0$}.
  Leftmost: the \emph{only} difference between the rollouts---the wrist input, with our $32$\,px patch at the bottom-center over the gripper; remaining columns: the third-person view over time (left\,$\rightarrow$\,right).
  Without the patch (top, \textcolor[HTML]{2E7D32}{green}) the policy completes the task; with it (bottom, \textcolor[HTML]{C62828}{red}) the same frozen policy is driven off course and fails.
}
\label{fig:qualitative}
\vspace{-7mm}
\end{figure*}
\section{Experiments}
\label{sec:experiments}

\subsection{Experimental Setup}
\label{sec:setup}

\subsubsection{Victim models.}
Our primary victim is $\pi_0$~\cite{black2024pi_0}, a flow-matching VLA that produces action chunks by integrating a learned denoising velocity field over $K{=}10$ Euler steps.
We additionally attack its successor $\pi_{0.5}$~\cite{black2025pi05}. 
Both policies use the publicly released LIBERO\cite{liu2023libero}-finetuned checkpoints and are kept entirely frozen: the attacker observes gradients to craft the patch but never updates the model weights.
\vspace{-2.5mm}
\subsubsection{Benchmark and protocol.}
We evaluate on the LIBERO benchmark~\cite{liu2023libero} across its four suites---\textbf{Spatial}, \textbf{Goal}, \textbf{Object}, and \textbf{Long} (LIBERO-10)---each containing 10 manipulation tasks. 
For every task we run 10 trials from the benchmark's predefined initial states, giving 100 episodes per suite. 
A patch is optimized \emph{only} on \texttt{libero\_spatial} observations and then evaluated unchanged on all four suites; \textbf{Spatial} is therefore in-distribution, while \textbf{Goal}, \textbf{Object}, and \textbf{Long} measure how well a single universal patch generalizes out of distribution.
\vspace{-2.5mm}
\subsubsection{Metrics.}
  For the no-patch reference we report the clean task success rate (TSR, \%). 
  For every attack we report the relative attack success rate $\mathrm{ASR}=(\mathrm{TSR}_\mathrm{clean}-\mathrm{TSR}_\mathrm{attack})/\mathrm{TSR}_\mathrm{clean}$---the fraction of originally-solvable tasks the attack breaks---so that an attack is not credited for tasks that fail even without perturbation; a higher ASR indicates a stronger attack.
  To complement this \emph{task-level} metric with an \emph{action-level} one, we additionally report the Normalized Action Discrepancy (NAD)~\cite{wang2025exploring}, which measures how far the attacked action deviates from a reference action, normalized per degree of freedom by its maximum possible deviation:
  \begin{equation} 
    \mathrm{NAD}=\frac{1}{I}\sum_{i=1}^{I}\frac{|a_i-a_i^{\mathrm{ref}}|}{\max\!\big(|a_i^{\mathrm{ref}}-a_{\min}|,\ |a_i^{\mathrm{ref}}-a_{\max}|\big)} .
  \end{equation}
    Unlike ASR, NAD captures how strongly a patch perturbs the action regardless of whether that perturbation actually breaks the task, so the two metrics are complementary.
    Since closed-loop rollouts provide no per-step ground-truth action, we take the reference $a^{\mathrm{ref}}$ to be the action the unperturbed policy produces on the same observation, in place of the dataset ground-truth used by~\cite{wang2025exploring}. NAD then measures how far the patch drives the action away from the clean policy's behavior.

\subsubsection{Baselines.}
We compare our single-step velocity attack against three baselines under the \emph{same} threat model and patch-size budget: (i) a random untrained patch, which isolates the effect of mere occlusion from that of optimization; (ii) UADA~\cite{wang2025exploring}, a universal patch optimized in the final action space; and (iii) EDPA~\cite{xu2025model}, which disrupts the vision-encoder embedding space. 
The sole exception is EDPA, which is ineffective at 32\,px size and is therefore reported at an effective size of 64\,px.
\vspace{-3mm}
\subsection{Implementation Details}
\label{sec:impl}

\subsubsection{Patch optimization.}
Unless otherwise noted, each patch is a $32\times32$ sticker (${\approx}2\%$ of the $224\times224$ wrist image) placed at the bottom-center of the wrist camera, over the gripper.
We optimize the single-step DRIFT objective of Eq.~\eqref{eq:single_step} at $k{=}0$ with PGD for 500 iterations at step size $\alpha{=}0.01$, starting from a uniform gray patch and clipping to the valid pixel range after every step. 
To make the patch universal, each iteration draws its conditioning observations from a pool of $696$ LIBERO wrist-camera frames rather than a single scene.
Because the objective targets only the first denoising step, each iteration requires a single forward and backward pass through the velocity field; a patch is optimized on a single NVIDIA RTX 3090 with 24GB VRAM.

\subsubsection{Evaluation.}
At test time, the optimized patch is pasted onto the wrist image at the same bottom-center location before each policy query, while the proprioceptive state, and language instruction are left untouched. 
We use each policy's default $K{=}10$-step denoising and its native action-chunk execution, re-querying the policy every few steps as in the original LIBERO evaluation. 

\subsection{Main Results}
\label{sec:results}

Table~\ref{tab:main_results} reports the results against the flow-matching victims $\pi_0$ and $\pi_{0.5}$ across the four LIBERO suites.
On $\pi_0$, DRIFT breaks \emph{essentially all} originally-solvable tasks (nearly $100\%$ ASR on every suite), far exceeding the action-space (UADA, $13.2\%$) and embedding-space (EDPA, $25.3\%$) baselines under the same patch-size budget; notably, EDPA is ineffective at our $32$\,px size and is reported with a larger $64$\,px patch, underscoring that embedding-space attacks require a substantially larger footprint. DRIFT also attains the highest NAD, confirming that it perturbs the action most strongly at the action level, not only in task outcome. The attack also generalizes to $\pi_{0.5}$, which appears more robust, requiring a larger $64$\,px patch to be broken.
Fig.~\ref{fig:qualitative} shows a representative success/failure pair, and Fig.~\ref{fig:traj} confirms the same effect at the level of the end-effector trajectory across five tasks.

\begin{table*}[t]
  \centering
  \caption{
      Results on four LIBERO suites against $\pi_0$ and $\pi_{0.5}$.
      \emph{No patch} reports clean TSR (\%); otherwise, ASR (\%, higher\,=\,stronger) and NAD (\%, higher\,=\,larger action discrepancy) are reported for each suite.
      Avg ASR and Avg NAD denote averages across the four suites.
      Patches are trained on \texttt{libero\_spatial} (in-distribution; Goal/Object/Long are out-of-distribution).
      $^{\dagger}$EDPA fails at 32\,px and is reported at 64\,px.
      Per-suite ASR and NAD values are reported as mean\,$\pm$\,std over 3 seeds; Avg columns report the corresponding averages across suites.
    }
  \label{tab:main_results}
  \setlength{\tabcolsep}{3.5pt}
  \small

  \resizebox{\textwidth}{!}{%
  \begin{tabular}{llcccccccccc}
  \toprule
  & & \multicolumn{2}{c}{\textbf{Spatial}} 
    & \multicolumn{2}{c}{\textbf{Goal}}
    & \multicolumn{2}{c}{\textbf{Object}}
    & \multicolumn{2}{c}{\textbf{Long}}
    & \multicolumn{2}{c}{\textbf{Avg}} \\
  \cmidrule(lr){3-4}
  \cmidrule(lr){5-6}
  \cmidrule(lr){7-8}
  \cmidrule(lr){9-10}
  \cmidrule(lr){11-12}
    \textbf{Victim} & \textbf{Method}
    & {\scriptsize\textbf{ASR} (\%)$\uparrow$} & {\scriptsize\textbf{NAD} (\%)$\uparrow$}
    & {\scriptsize\textbf{ASR} (\%)$\uparrow$} & {\scriptsize\textbf{NAD} (\%)$\uparrow$}
    & {\scriptsize\textbf{ASR} (\%)$\uparrow$} & {\scriptsize\textbf{NAD} (\%)$\uparrow$}
    & {\scriptsize\textbf{ASR} (\%)$\uparrow$} & {\scriptsize\textbf{NAD} (\%)$\uparrow$}
    & {\scriptsize\textbf{ASR} (\%)$\uparrow$} & {\scriptsize\textbf{NAD} (\%)$\uparrow$} \\
  \midrule

  $\pi_0$~\cite{black2024pi_0}
    & No patch \textit{(TSR)}
    & 98.0 & -- & 97.0 & -- & 98.0 & -- & 87.0 & -- & 95.0 & -- \\

    & Random patch {\small(32\,px)}
    & 0.7$\pm$1.3 & 4.3$\pm$0.1
    & 5.2$\pm$1.5 & 3.9$\pm$0.0
    & 0.3$\pm$0.5 & 2.7$\pm$0.0
    & 8.8$\pm$7.0 & 3.4$\pm$0.1
    & 3.7 & 3.6 \\

    & UADA~\cite{wang2025exploring} {\small(32\,px)}
    & 11.6$\pm$15.7 & 5.2$\pm$1.0
    & 12.0$\pm$8.3 & 4.5$\pm$0.8
    & 7.5$\pm$6.5 & 3.3$\pm$0.8
    & 21.5$\pm$7.6 & 4.0$\pm$0.7
    & 13.2 & 4.3 \\

    & EDPA~\cite{xu2025model}$^{\dagger}$ {\small(64\,px)}
    & 23.1$\pm$16.4 & 6.7$\pm$1.7
    & 28.2$\pm$15.1 & 5.9$\pm$1.4
    & 18.7$\pm$12.9 & 4.0$\pm$0.9
    & 31.0$\pm$20.6 & 5.0$\pm$0.9
    & 25.3 & 5.4 \\

  \cmidrule(lr){2-12}
    & \textbf{DRIFT} {\small(32\,px)}
    & \textbf{100.0$\pm$0.0} & \textbf{10.5$\pm$2.4}
    & \textbf{99.7$\pm$0.5} & \textbf{8.5$\pm$2.0}
    & \textbf{100.0$\pm$0.0} & \textbf{7.5$\pm$2.0}
    & \textbf{99.6$\pm$0.5} & \textbf{7.4$\pm$1.6}
    & \textbf{99.8} & \textbf{8.5} \\

  \midrule

  $\pi_{0.5}$~\cite{black2025pi05}
    & No patch \textit{(TSR)}
    & 100.0 & -- & 100.0 & -- & 98.0 & -- & 91.0 & -- & 97.3 & -- \\

    & Random patch {\small(64\,px)}
    & 4.3$\pm$1.2 & 2.4$\pm$0.0
    & 7.7$\pm$1.7 & 3.4$\pm$0.1
    & 2.7$\pm$0.5 & 2.6$\pm$0.0
    & 11.4$\pm$1.4 & 2.5$\pm$0.0
    & 6.5 & 2.7 \\

    & UADA~\cite{wang2025exploring} {\small(64\,px)}
    & 34.7$\pm$46.2 & 4.0$\pm$2.9
    & 23.3$\pm$30.2 & 4.6$\pm$3.3
    & 33.0$\pm$47.4 & 4.0$\pm$3.4
    & 31.1$\pm$40.1 & 3.1$\pm$1.8
    & 30.5 & 3.9 \\

    & EDPA~\cite{xu2025model}$^{\dagger}$ {\small(64\,px)}
    & 0.3$\pm$0.5 & 2.3$\pm$0.1
    & 3.7$\pm$0.5 & 2.8$\pm$0.1
    & 1.4$\pm$0.5 & 1.7$\pm$0.1
    & 0.7$\pm$2.9 & 1.8$\pm$0.1
    & 1.5 & 2.1 \\

  \cmidrule(lr){2-12}
    & \textbf{DRIFT} {\small(64\,px)}
    & \textbf{100.0$\pm$0.0} & \textbf{23.5$\pm$2.0}
    & \textbf{97.3$\pm$3.8} & \textbf{21.4$\pm$3.3}
    & \textbf{100.0$\pm$0.0} & \textbf{18.0$\pm$0.8}
    & \textbf{100.0$\pm$0.0} & \textbf{17.3$\pm$1.0}
    & \textbf{99.3} & \textbf{20.0} \\

  \bottomrule
  \end{tabular}%
  }
\end{table*}

\subsubsection{Failure mode: phantom grasp.}
\begin{figure*}[t]
\centering
\begin{subfigure}{0.15\textwidth}\centering\includegraphics[width=\linewidth]{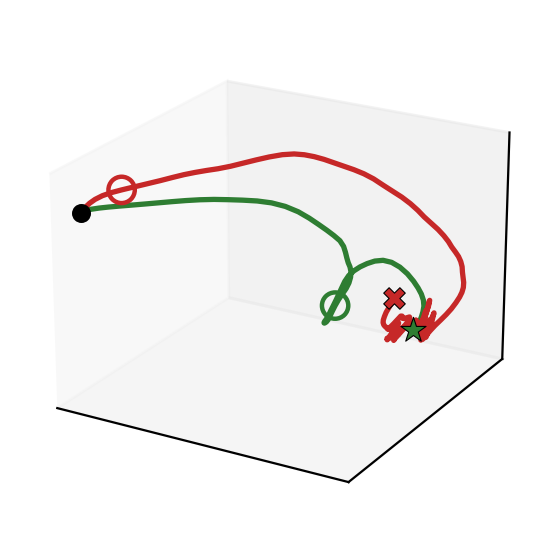}\caption{}\label{fig:traj_a}\end{subfigure}\hfill
\begin{subfigure}{0.15\textwidth}\centering\includegraphics[width=\linewidth]{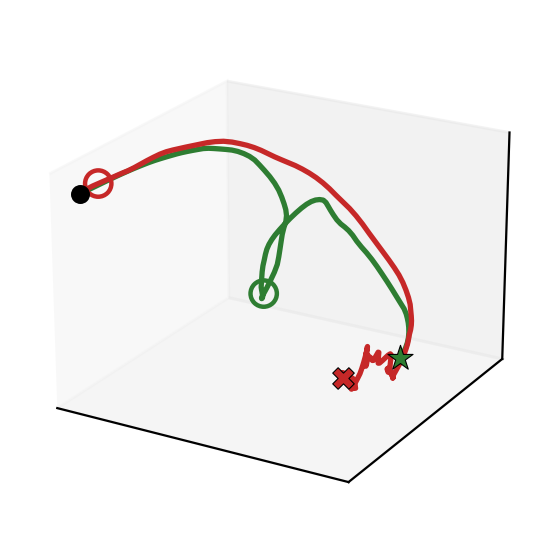}\caption{}\label{fig:traj_b}\end{subfigure}\hfill
\begin{subfigure}{0.15\textwidth}\centering\includegraphics[width=\linewidth]{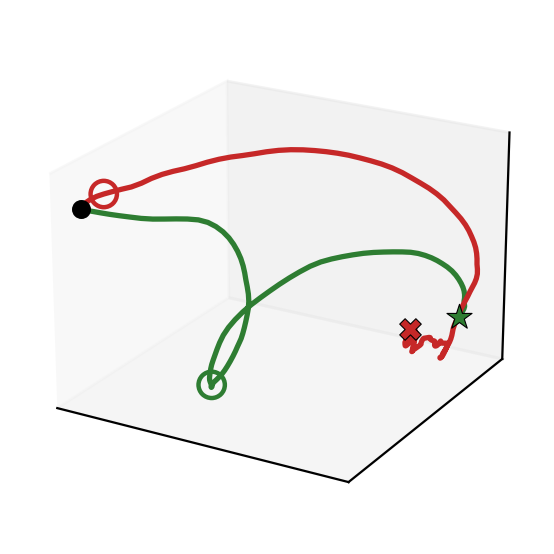}\caption{}\label{fig:traj_c}\end{subfigure}\hfill
\begin{subfigure}{0.15\textwidth}\centering\includegraphics[width=\linewidth]{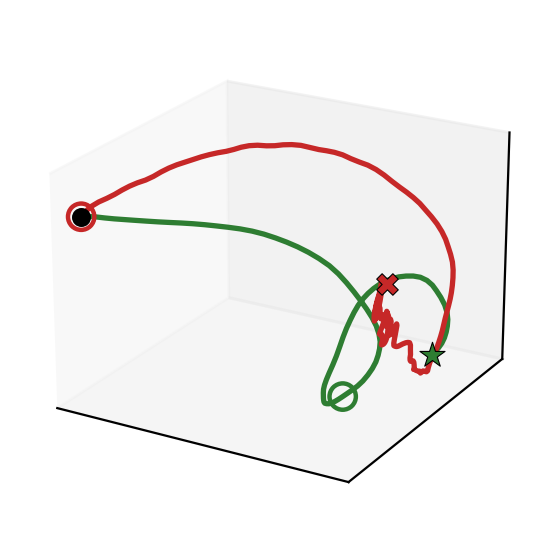}\caption{}\label{fig:traj_d}\end{subfigure}\hfill
\begin{subfigure}{0.15\textwidth}\centering\includegraphics[width=\linewidth]{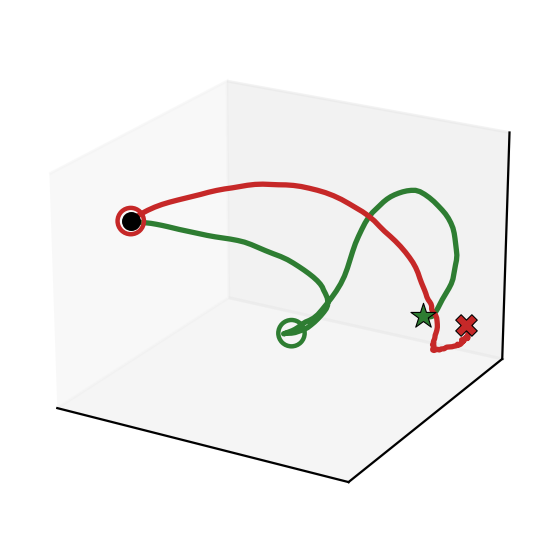}\caption{}\label{fig:traj_e}\end{subfigure}
\caption{
  \textbf{End-effector trajectories across five LIBERO-Spatial tasks} (a--e), each from an identical initial state.
  \textcolor[HTML]{2E7D32}{Green}\,=\,clean (succeeds), \textcolor[HTML]{C62828}{red}\,=\,patched (fails); both share the start ($\bullet$), with $\star$/$\times$ the successful/failed end and the hollow circle the first gripper closing.
  The clean policy reaches and grasps along a short, direct path, whereas the patched policy is consistently diverted and never completes the grasp.
}

\label{fig:traj}
\end{figure*}

The trajectories in Fig.~\ref{fig:traj} expose a consistent mechanism behind the near-total failure.
Under the patch, the policy commands the gripper to close almost immediately---within the first $7$ control steps in all five episodes (mean $3.8$)---long before the arm has approached any object.
The clean policy, by contrast, keeps the gripper open until it reaches the bowl, closing only around step $46$ on average (\textcolor[HTML]{2E7D32}{green} hollow circles in Fig.~\ref{fig:traj}).
The failure is spatial as well: the patched arm keeps moving (mean displacement $40.6\,\mathrm{cm}$, exceeding the clean policy's) yet never approaches the object, coming no closer than $18.6\,\mathrm{cm}$ on average.
The attacked robot thus advances through the scene already clenched on nothing---a \emph{phantom grasp}---and never executes the reach-then-grasp sequence the task demands.
This degenerate, grasp-locked behavior emerges across all tasks rather than steering the arm toward any particular target pose, consistent with our untargeted threat model.

\subsection{Further Analysis}
\label{sec:further}

\subsubsection{Perturbation budget.}
Constraining the patch to an $\ell_\infty$ ball of radius $\epsilon$ around the initialization and sweeping $\epsilon$ reveals a sharp \emph{phase transition} rather than a gradual trade-off (Fig.~\ref{fig:eps_a}, \textcolor[HTML]{C62828}{red}): ASR stays near zero for $\epsilon\!\le\!0.05$ ($\le\!8.2\%$), then jumps to $97\%$ at $\epsilon{=}0.1$---essentially all-or-nothing: below a perturbation budget threshold, the patch cannot corrupt the denoising trajectory through the first step, while above it, the cascade effect induces task failure.

\subsubsection{Color vs. grayscale patch}
\begin{figure}[t]
\centering
\vspace{-3mm}
\begin{subfigure}[c]{0.5\linewidth}\centering\includegraphics[width=\linewidth]{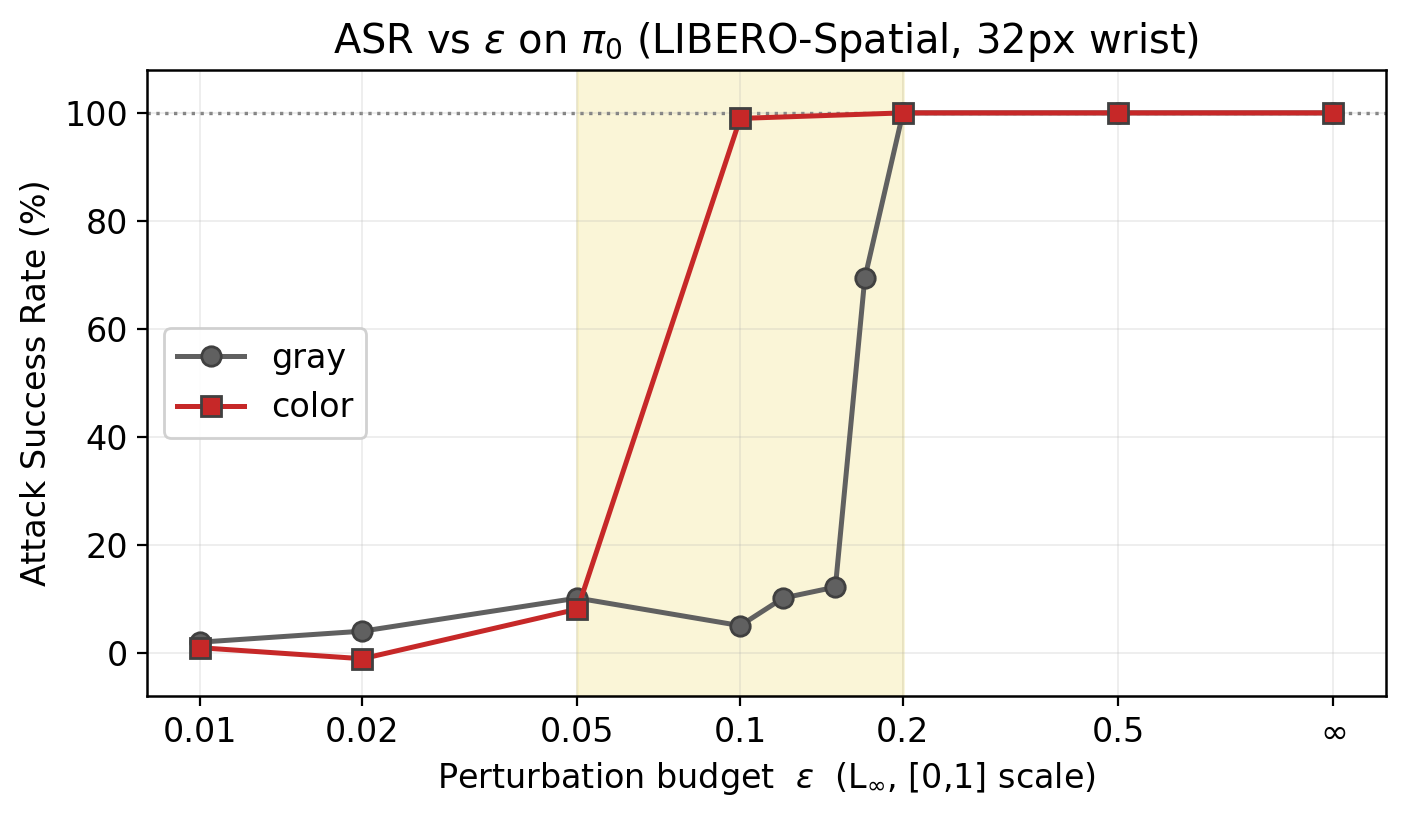}\caption{}\label{fig:eps_a}\end{subfigure}\hfill
\begin{subfigure}[c]{0.5\linewidth}\centering\includegraphics[width=\linewidth]{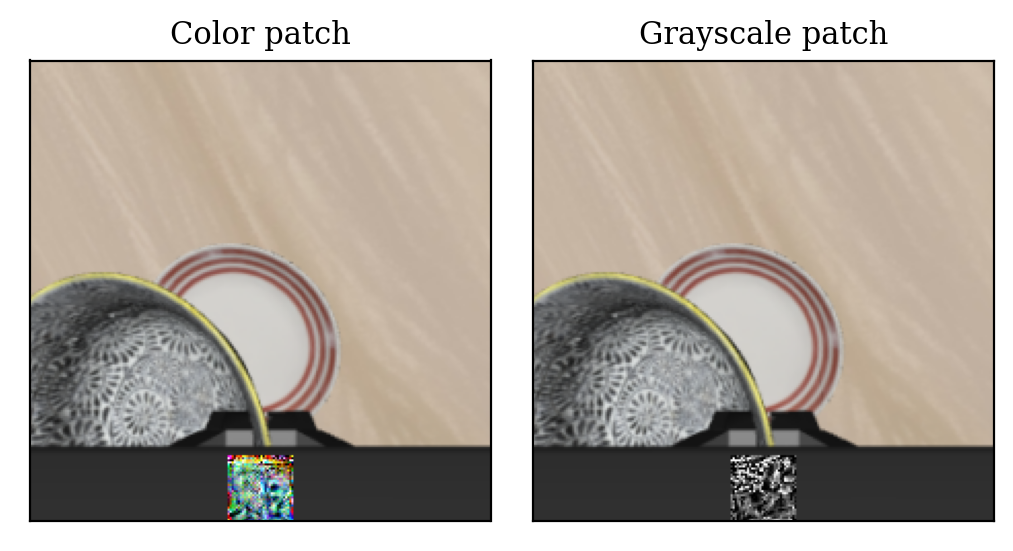}\caption{}\label{fig:eps_b}\end{subfigure}
\caption{\textbf{Perturbation budget and stealthiness}
(a)~Attack success vs.\ $\ell_\infty$ budget $\epsilon$: both color and grayscale patches show sharp phase transitions rather than gradual trade-off; using the color patches (\textcolor[HTML]{C62828}{red}) saturates at roughly half the budget of the grayscale patch (\textcolor[HTML]{808080}{gray})
(b)~Our attack does not need to be visually salient: applied to the same wrist view, the default patch (left) is conspicuous against the gripper, whereas a grayscale-constrained patch (right) blends into the dark gripper region---the perturbation can be realized far more inconspicuously.}
\vspace{-3mm}
\label{fig:eps}
\end{figure}

We additionally test the case where the adversarial patch is restricted to grayscale, which enhances the stealthiness of the patch. 
Fig.~\ref{fig:eps_a} confirms that the grayscale patch is also effective in attacking, reaching ASR 100\% when a sufficient perturbation budget is allowed, while enhancing stealthiness in Fig.~\ref{fig:eps_b}.
As expected, the grayscale patch requires about two times the perturbation budget needed for the color patch for successful attack.

\subsubsection{Cross-model transfer.}
We test whether a DRIFT patch optimized against one flow-matching victim transfers to the other without re-optimization (Table~\ref{tab:transfer}). Consistent with our main results (Table~\ref{tab:main_results}), $\pi_{0.5}$ appears more robust: a patch ports more readily onto $\pi_0$ ($24.7\%$) than onto $\pi_{0.5}$ ($8.9\%$). We conclude that the early-step vulnerability is a \emph{shared} property of flow-matching VLAs---both are breakable---but the specific perturbation exploiting it is \emph{model-specific}. This reinforces our white-box threat model~(Sec.~\ref{sec:threat_model}): effective attacks need gradient access to the deployed policy and cannot simply be ported across model versions.

\begin{table}[t]
\centering
\vspace{3mm}
\caption{\textbf{Cross-model transfer} (ASR \%). A DRIFT patch
optimized on the \emph{source} victim is applied \emph{without re-optimization} to the \emph{target}. White-box (in-model) ASR is nearly $100\%$ for both victims (Table~\ref{tab:main_results});
all rows are cross-model. Raising the patch from $32{\to}64$\,px lifts $\pi_0$ to $\pi_{0.5}$
transfer only $0.4{\to}8.9\%$, so the weak transfer reflects \emph{model-specific} patches, not a size artifact.}
\label{tab:transfer}
\setlength{\tabcolsep}{6pt}
\begin{tabular}{llccccc}
\toprule
\textbf{Transfer} & \textbf{Patch} & \textbf{Spatial} & \textbf{Goal} & \textbf{Object} & \textbf{Long} & \textbf{Avg} \\
\midrule
$\pi_0 \!\to\! \pi_{0.5}$ & 32\,px & 4.0 & 1.0 & -1.0 & -2.2 & 0.4 \\
$\pi_0 \!\to\! \pi_{0.5}$ & 64\,px & 5.0 & 6.0 & 9.2 & 15.4 & \textbf{8.9} \\
$\pi_{0.5} \!\to\! \pi_0$ & 64\,px & 19.4 & 23.7 & 14.3 & 41.4 & \textbf{24.7} \\
\bottomrule
\end{tabular}
\end{table}

\section{Conclusion}
\label{sec:conclusion}

We introduced \textbf{DRIFT}, a test-time universal adversarial patch that corrupts the denoising velocity field of a \emph{frozen} flow-matching VLA through a single sticker on the robot's gripper. Our central finding is that the \emph{first} denoising step is all that must be perturbed: attacking only the first step is stronger and cheaper than a wider window---a ``less-is-more'' effect we trace to a gradient conflict that runs opposite to the training-time backdoor regime, where a wide early-step window is instead required. DRIFT breaks essentially all $\pi_0$ tasks with a single, small patch, far exceeding action- and embedding-space baselines, and generalizes to the more robust $\pi_{0.5}$. The reported robustness of flow-matching VLAs is thus largely an artifact of attacks that ignore the denoising ODE, calling for defenses that protect its earliest steps.

\section{Acknowledgements}
\label{sec:acknowledgements}
This work was supported by the National Research Foundation of Korea (NRF) grant (No. RS-2024-00453301, RS-2025-00517159) and by Institute of Information communications Technology Planning Evaluation (IITP) grant (IITP-2025-RS-2024-00428780).


%
%
\bibliographystyle{splncs04}
\bibliography{main}
\clearpage
\appendix
\begin{center}
    {{\Large \textbf{DRIFT: Derailing Denoising Trajectories of Flow-Matching VLAs with Adversarial Patch Attack}\\[0.8em]
    Supplementary Materials}}
\end{center}

This supplementary material includes:
\begin{itemize}
    \item $8$ supplementary wrist-camera rollout videos (in the accompanying zip):
    one clean and one DRIFT-patched pair per LIBERO suite (Spatial, Goal, Object,
    Long), each pair showing the \emph{same} task so the clean (successful) and
    patched (failed) rollouts can be directly compared.
    \item Sec.~\ref{sec:supp_setting}: Experimental Setting Details
    \item Sec.~\ref{sec:supp_pcgrad}: Further analysis on gradient conflict
    \item Sec.~\ref{sec:supp_saliency}: Attention Map on the Target Object
    \item Sec.~\ref{sec:supp_defense}: A Simple Defense Baseline (JPEG Compression)
\end{itemize}

\section{Experimental Setting Details}
\label{sec:supp_setting}

\subsection{LIBERO Evaluation Protocol}
Each policy observes RGB images natively rendered at $256\times256$ and resized with
zero-padding to $224\times224$ before being passed to the network, matching the
policy's training-time preprocessing.
At the start of each episode we execute $10$ warm-up steps (no policy query) to let
the simulation settle, following the standard LIBERO evaluation protocol.
Both $\pi_0$ and $\pi_{0.5}$ predict an action chunk of which we execute the first
$5$ actions open-loop before re-querying the policy (\emph{replan} every $5$ steps).
This amounts to $4\times100=400$ episodes per attack/model/seed configuration
(Sec.~\ref{sec:setup}, main paper).

\subsection{LIBERO-Spatial wrist-frame collection}
We roll out the base policy on all 10 tasks (3 episodes/task; separated from evaluation episodes), storing frames at each action-generation step (696 total, resize to $224*224$).

\subsection{Patch Placement}
For a patch of height $h$ and width $w$ placed at the wrist image's
bottom-center (Sec.~\ref{sec:impl}, main paper), the top-left corner falls at pixel
$(y,x) = (224-p_h,\ (224-p_w)/2)$. Table~\ref{tab:supp_patch_coords} gives the
resulting coordinates for the two patch sizes used in this paper.

\begin{table}[h]
\centering
\caption{Exact patch placement coordinates (top-left corner, pixels).}
\label{tab:supp_patch_coords}
\begin{tabular}{lcc}
\toprule
\textbf{Victim} & \textbf{Patch size} & \textbf{$(y,x)$} \\
\midrule
$\pi_0$    & $32\times32$ & $(192,\ 96)$ \\
$\pi_{0.5}$ & $64\times64$ & $(160,\ 80)$ \\
\bottomrule
\end{tabular}
\end{table}

\subsection{Baseline Reimplementation}
All baselines share DRIFT's optimization budget (PGD, 500 iterations, step size
$\alpha{=}0.01$, same universal observation pool) so that differences in ASR reflect
the attack objective rather than the optimization procedure.

\noindent\textbf{Random patch.} A single patch of uniform random pixel values is
sampled once (per seed) and left unoptimized, isolating the effect of mere occlusion
from that of gradient-based optimization.

\noindent\textbf{UADA~\cite{wang2025exploring}.} We maximize the squared deviation
between the final denoised action chunks obtained under the clean and patched
observations, $\|\mathbf{A}^{1}_{\mathrm{adv}}-\mathbf{A}^{1}_{\mathrm{clean}}\|_2^2$,
backpropagating through the full $K$-step rollout (i.e., an action-space objective
with no explicit per-step velocity term).

\noindent\textbf{EDPA~\cite{xu2025model}.} We disrupt the vision-encoder embedding
space with two terms computed on the frozen PaliGemma vision/language tower: (i) an
InfoNCE term (temperature $0.07$) that pushes each patched patch-embedding away from
its clean counterpart, and (ii) an image--language alignment term that maximizes the
change in cosine similarity between patch embeddings and language-token embeddings.
The two terms are combined with an EMA-balanced weighted sum (decay $0.9$, mixing
weight $0.5$) following the official formulation, and require no denoising rollout.

\section{Further analysis on gradient conflict}
\label{sec:supp_pcgrad}
\begin{table}[h]
\centering
\caption{PCGrad resolves the conflict and restores the attack}
\label{tab:pcgrad}
\begin{tabular}{cccc}
\toprule
$M$ & $R$: before$\to$after & $\Delta R$ & ASR: sum$\to$PCGrad \\
\midrule
3  & $.94\to.99$ & $.05$  & $92\to100$ \\
5  & $.88\to.97$ & $.09$  & $77\to100$ \\
10 & $.45\to.74$ & $+.29$ & $14.9\to30.9$ \\
\bottomrule
\end{tabular}
\end{table}

To further verify that the degradation of multi-step optimization stems from per-step gradient conflict problem, we explicitly mitigate gradient conflict using PCGrad~\cite{yu2020gradient}. 
Table~\ref{tab:pcgrad} compares the original gradient summation with PCGrad for different numbers of targeted steps $M$. Resolving gradient conflicts consistently improves survival ratio $R$, with the effect becoming more pronounced as more steps are included. 
These results provide additional evidence that the ``less-is-more’’ behavior observed in multi-step optimization is largely driven by gradient conflict: as increasingly misaligned step-wise objectives are jointly optimized, their gradients partially cancel in input space, weakening the resulting adversarial patch. Explicitly resolving these conflicts recovers a substantial portion of the lost attack strength, although severe conflicts over long optimization windows cannot be fully eliminated.
\section{Attention Map on the Target Object}
\label{sec:supp_saliency}

\begin{figure}[t]
\centering
\includegraphics[width=0.85\linewidth]{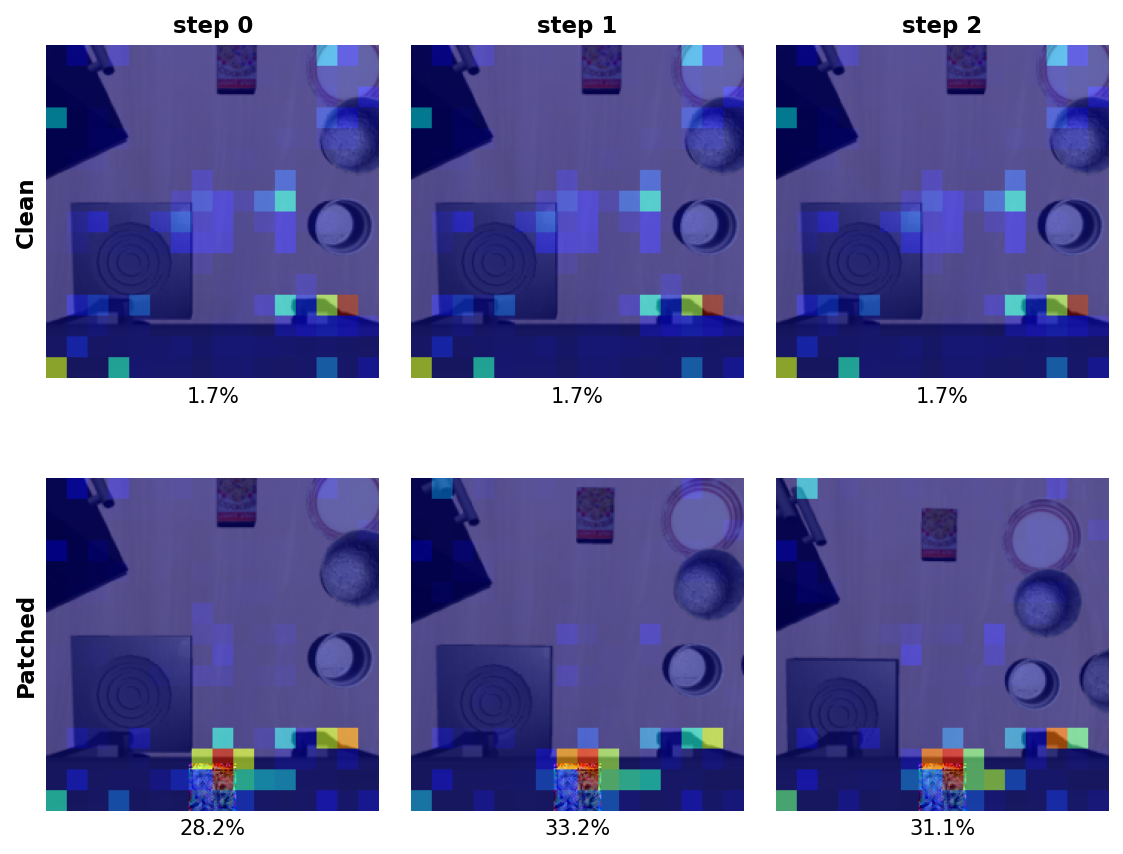}
\caption{\textbf{The patch absorbs a disproportionate share of the object-referring
token's attention.} ``bowl'' token attention onto the wrist image at three early
steps. Top row (Clean): the unpatched rollout of the same task, which
succeeds. Bottom row (Patched): the
DRIFT-patched rollout, which fails. The percentage beneath each panel is the fraction
of the ``bowl'' token's attention falling inside the $32{\times}32$ patch region
(chance level $3.5\%$). Without the patch this region stays below chance ($1.7\%$);
once patched, $28$--$33\%$ of the attention concentrates there.}
\label{fig:supp_attention_patch}
\end{figure}

The phantom-grasp failure mode (main paper, Sec.~\ref{sec:results}) shows that the
patched policy commands the gripper to close almost immediately, without ever
approaching the object. Here we test a complementary hypothesis at the level of the
policy's internal mechanism: does the patch redirect the model's \emph{attention} away
from the object to be grasped? We extract the policy's actual softmax self-attention
weights to answer this directly.

\subsubsection{Method.}
We analyze the LIBERO-Spatial task \emph{``pick up the black bowl between the plate
and the ramekin and place it on the plate''} (task~0, the same task shown in
Fig.~\ref{fig:traj}a of the main paper). $\pi_0$'s prefix (all image and language
tokens) attends to itself through a stack of self-attention layers; we extract the
real softmax attention weights for the language token corresponding to the object
noun (``bowl''), averaged over layers and heads, restricted to the $16{\times}16{=}256$
visual tokens of a given camera.

\subsubsection{Patch Redirects Attention}
We ask a narrow, well-defined
question: does the ``bowl'' token's attention onto the
$32{\times}32$ patch region change when the patch is present, relative to a chance
level set purely by that region's area fraction ($3.5\%$ of the $256$ tokens)? Under
the clean observation this region receives \emph{below}-chance attention ($1.7\%$).
Fig.~\ref{fig:supp_attention_patch} (bottom) shows that across the first three steps of the failed rollout, the patch region absorbs $28$--$33\%$ of the ``bowl'' token's
attention ---an $8$--$9\times$ increase over chance,
concentrated precisely on a region that is otherwise task-irrelevant.

\section{A Simple Defense Baseline: JPEG Compression}
\label{sec:supp_defense}

\begin{figure}[t]
\centering
\includegraphics[width=0.72\linewidth]{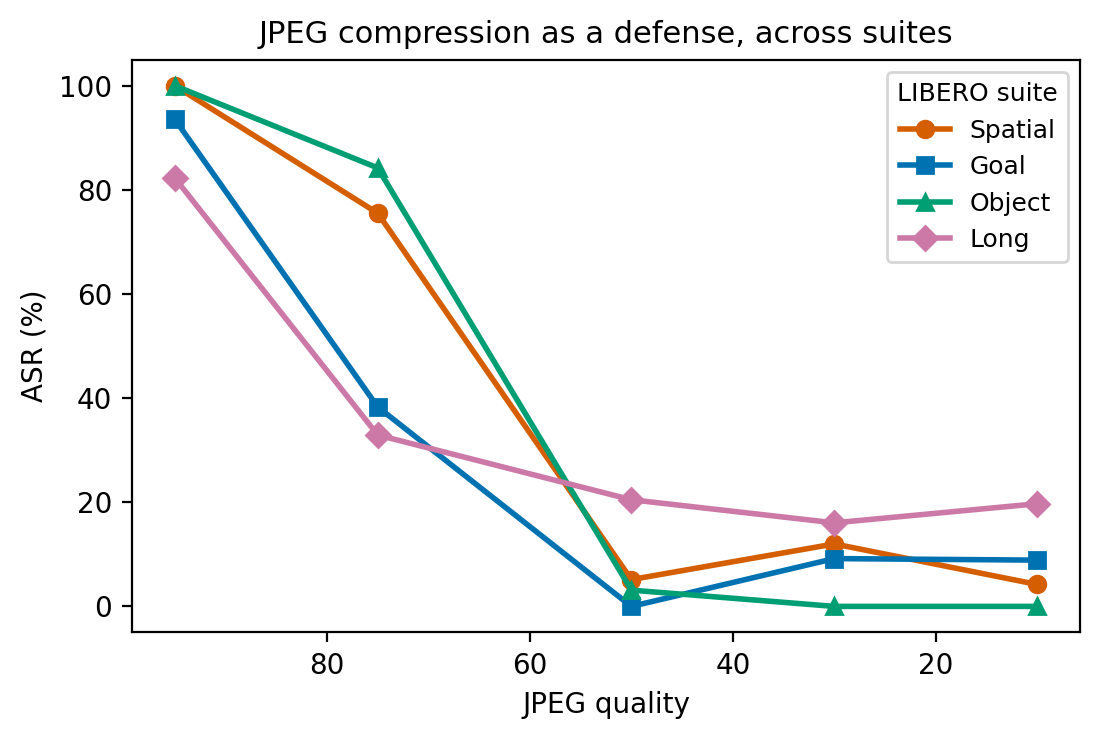}
\caption{\textbf{JPEG compression as a defense, across LIBERO suites.} ASR vs.\
JPEG quality. ASR falls off progressively as $Q$ decreases; once compression is
strong enough (roughly $Q\!\lesssim\!50$--$60$) the attack is largely blocked on
Spatial, Goal, and Object (near-zero ASR). The Long suite is an exception, plateauing
at a markedly higher ASR ($16$--$21\%$) even under the strongest compression tested.}
\label{fig:supp_jpeg_defense}
\end{figure}
As a first, training-free defense probe, we test whether the JPEG compression-based defense method~\cite{dziugaite2016study, das2017keeping} can mitigate DRIFT. Across all four LIBERO suites, we sweep
the JPEG quality factor $Q\in\{95,75,50,30,10\}$ and re-measure both the
clean task success rate and the DRIFT-patched task success rate under each $Q$, computing
$\mathrm{ASR}(Q)=(\mathrm{TSR}_{\mathrm{clean}}(Q)-\mathrm{TSR}_{\mathrm{drift}}(Q))/\mathrm{TSR}_{\mathrm{clean}}(Q)$ (100 episodes each);
clean TSR stayed in the $71$--$100\%$ range throughout, so the ASR drops are
not an artifact of the defense simply breaking the policy.

Fig.~\ref{fig:supp_jpeg_defense} shows that ASR falls off progressively as the JPEG
quality $Q$ decreases. At mild compression ($Q{\geq}75$) DRIFT remains largely
effective on Spatial and Object ($75$--$100\%$ ASR), and is already weaker and more
variable on Goal ($38$--$94\%$). As $Q$ drops further the attack weakens, and once
compression is strong enough---roughly $Q\!\lesssim\!50$--$60$---it is largely
blocked on Spatial, Goal, and Object, with ASR down to near zero ($0$--$12\%$). The
Long suite is an exception: ASR only falls to $16$--$21\%$ and plateaus there even at
the strongest compression tested ($Q{=}10$), roughly $4\times$ higher than the other
three suites at the same setting.

We further evaluate JPEG compression against an \emph{adaptive} attacker that
incorporates JPEG compression directly into the PGD optimization loop. At
$Q{=}50$, where the non-adaptive attack is largely suppressed, the adaptive attack
recovers the ASR from $2\%$ to $28.7\%$. This recovery shows that JPEG compression
does not provide complete robustness once the attacker accounts for the defense.
Nevertheless, the attack remains substantially weaker than without the defense,
suggesting that JPEG compression can serve as a simple, training-free mitigation,
rather than a general-purpose defense against adaptive attacks.
\end{document}